\documentclass{article}
\usepackage{spconf,amsmath,graphicx}
\usepackage{tabularx}
\usepackage{tablefootnote}
\usepackage{float}

\usepackage{enumitem}
\setlist{nosep, leftmargin=14pt}

\usepackage{mwe} 

\title{Cross-Dataset Generalization For Retinal Lesions Segmentation}
\name{Cl\'ement Playout, Farida Cheriet}
\address{LIV4D, \'Ecole Polytechnique de Montr\'eal, Canada}
\begin{document}
	%
	\maketitle
	\begin{abstract}
		Identifying lesions in fundus images is an important milestone toward an automated and interpretable diagnosis of retinal diseases. To support research in this direction, multiple datasets have been released, proposing groundtruth maps for different lesions. However, important discrepancies exist between the annotations and raise the question of generalization across datasets. This study characterizes several known datasets
		and compares different techniques that have been proposed to enhance the generalisation performance of a model,
		such as stochastic weight averaging, model soups and ensembles. Our results provide insights into how to combine coarsely labelled data with a finely-grained dataset in order to improve the lesions segmentation.
		
	\end{abstract}
	\section{Introduction}
	\label{sec:intro}
	
	The American Academy of Ophthalmology estimates that approximately one-third of Americans risk developing diabetes mellitus \cite{flaxelDiabeticRetinopathyPreferred2020}, an alarming estimation corroborated by studies in many countries \cite{lamWorldwideDiabetesEpidemic2012}. Among its potential complications, diabetic retinopathy (DR) is the leading
	cause of legal blindness in the working-age population worldwide.
	Effective screening strategies are needed to treat the disease, increasingly involving tele-screening initiatives \cite{kalogeropoulosRoleTeleophthalmologyDiabetic2020}.
	AI-based models have been actively researched to reduce costs, expand screening opportunities and scale to nation-wide populations. In parallel, the need for clinically interpretable models has led to developing methods to automatically identify lesions in the retina from fundus images. To do so, many datasets have been published in recent years to foster research on segmentation algorithms. However, given that labelling at a pixel level is a time-consuming and expensive task, strategies are required to mitigate these costs. For large datasets, adopting a coarser granularity for the manual labelling of lesions is often the only choice. This leads to important discrepancies in terms of raw images and labelling styles between datasets and naturally raises the question of cross-dataset generalization, the latter question being relevant in most applications of computer vision. 
	
	In the past decade, the field of segmentation has seen an almost complete shift from traditional computer vision algorithms to a new de-facto standard built on fully convolutional neural networks (CNNs). Numerous architectures have been designed to segment one or multiple lesion types relevant to the detection of DR \cite{yanLearningMutuallyLocalGlobal2019a, zhouBenchmarkStudyingDiabetic2021c, weiLearnSegmentRetinal2021a, guoLSegEndtoendUnified2019a}.
	Even with these modern architectures, training a deep learning model generally requires a large amount of data. Recently, multiple datasets have been released, usually with accompanying architectures for segmentation and grading. However, these datasets are not standardized, thus each has its specificities. The IDRID challenge \cite{porwalIndianDiabeticRetinopathy2018a} introduced a fundus dataset composed of 81 images with mask annotations for four lesion types.
	DDR was released \cite{liDiagnosticAssessmentDeep2019b} alongside a study on both segmentation and grading of DR. That work reported high accuracy results for DR grading but highlighted the difficulties posed by segmentation. More recently, the FGADR dataset has been released \cite{zhouBenchmarkStudyingDiabetic2021c}: with 1852 annotated images, it is to our knowledge the largest dataset for fundus segmentation. Their study proposes a thorough analysis of the benefits of segmentation masks for DR grading as well as for the detection of other pathologies. A similar work was proposed in \cite{weiLearnSegmentRetinal2021a}, introducing another large dataset (we refer to it as Retinal Lesions) as well as a new segmentation architecture called Lesion-Net.
	
	In this paper, we study the generalization ability of different segmentation models over these four datasets and include a fifth one composed of publicly available images annotated by a team of experts. We first characterize each dataset, then use them to analyse the generalization properties of different architectures. We then compare different approaches that have been proposed to improve a model's generalization: Stochastic Weight Averaging \cite{izmailovAveragingWeightsLeads2018}, Model soups \cite{wortsmanModelSoupsAveraging} and an ensemble of models. To our knowledge, neither of the first two have been experimented with segmentation. Our work provides different insights on how to combine fine and coarse datasets to improve a model's generalization ability. These findings are important, as the cost of groundtruth labelling 
	is a major obstacle to the creation of new datasets.

	\section{Methodology}
	\subsection{Datasets characterization}
	
	\begin{table}[h]
		\centering
		\small
		\begin{tabularx}{\columnwidth}{l|XX|r}
			Datasets &  Size & Lesions &  ID \\
			\hline
			
			IDRID \cite{porwalIndianDiabeticRetinopathy2018a} & 81 & \textbf{Ex CWS He Ma} & IDR \\
			DDR \cite{liDiagnosticAssessmentDeep2019b}& 757 & \textbf{Ex CWS He Ma} & DDR \\
			FGADR \cite{zhouBenchmarkStudyingDiabetic2021c} & 1842 & \textbf{Ex CWS He Ma} NV IRMA & FGA \\
			Retinal lesions \cite{weiLearnSegmentRetinal2021a} & 1593 & \textbf{Ex CWS He Ma} PrHE VHe & RET \\
			Messidor-MAPLES \cite{decenciere_feedback_2014, maples_dr} & 200 & \textbf{Ex CWS He Ma} & MES \\
			\hline
		\end{tabularx}       
		\caption{Summary of the datasets used in our study. In our work, we only considered the lesions shown in bold. \textit{Ex: Hard Exudate, CWS: Cotton Wool Spot, He: Hemorrhage, Ma:Microaneurysm, NV: Neo-vascularization, IRMA: Intraretinal Microvascular Abnormality, PrHe: Preretinal He, VHe: vitreous He}. In the rest of the paper, the datasets are referred by their three-letter ID. }\label{tab:Datasets}
	\end{table}
	
	Five datasets were used in our study as described in Table \ref{tab:Datasets}. The last one, Messidor-MAPLES-DR, was derived from the publicly available MESSIDOR dataset \cite{decenciere_feedback_2014}, composed of 1200 images. We asked a panel of seven experienced ophthalmologists to label a subset of 200 images using an online labelling tool developed in-house for this purpose (\cite{maples_dr}).

\begin{figure}[h]
		\centering
		\centerline{\includegraphics[width=\linewidth]{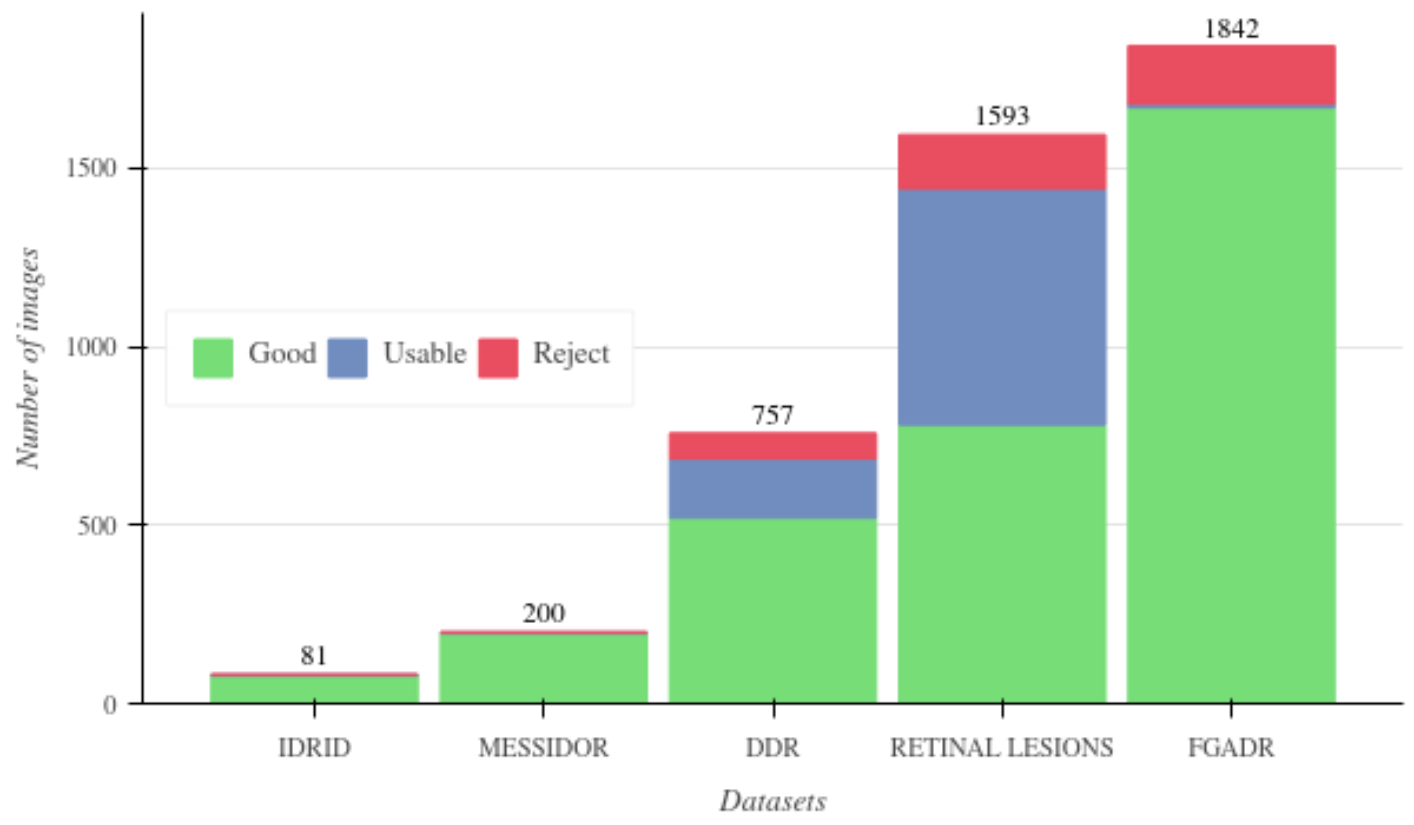}}
		\caption{Image quality as evaluated by the MCF-Net for each dataset. The distributions are normalized.}
		\label{fig:image_quality}
	\end{figure}

\begin{figure}
		\begin{minipage}[b]{.49\linewidth}
			\centering
			\centerline{\includegraphics[width=4.0cm]{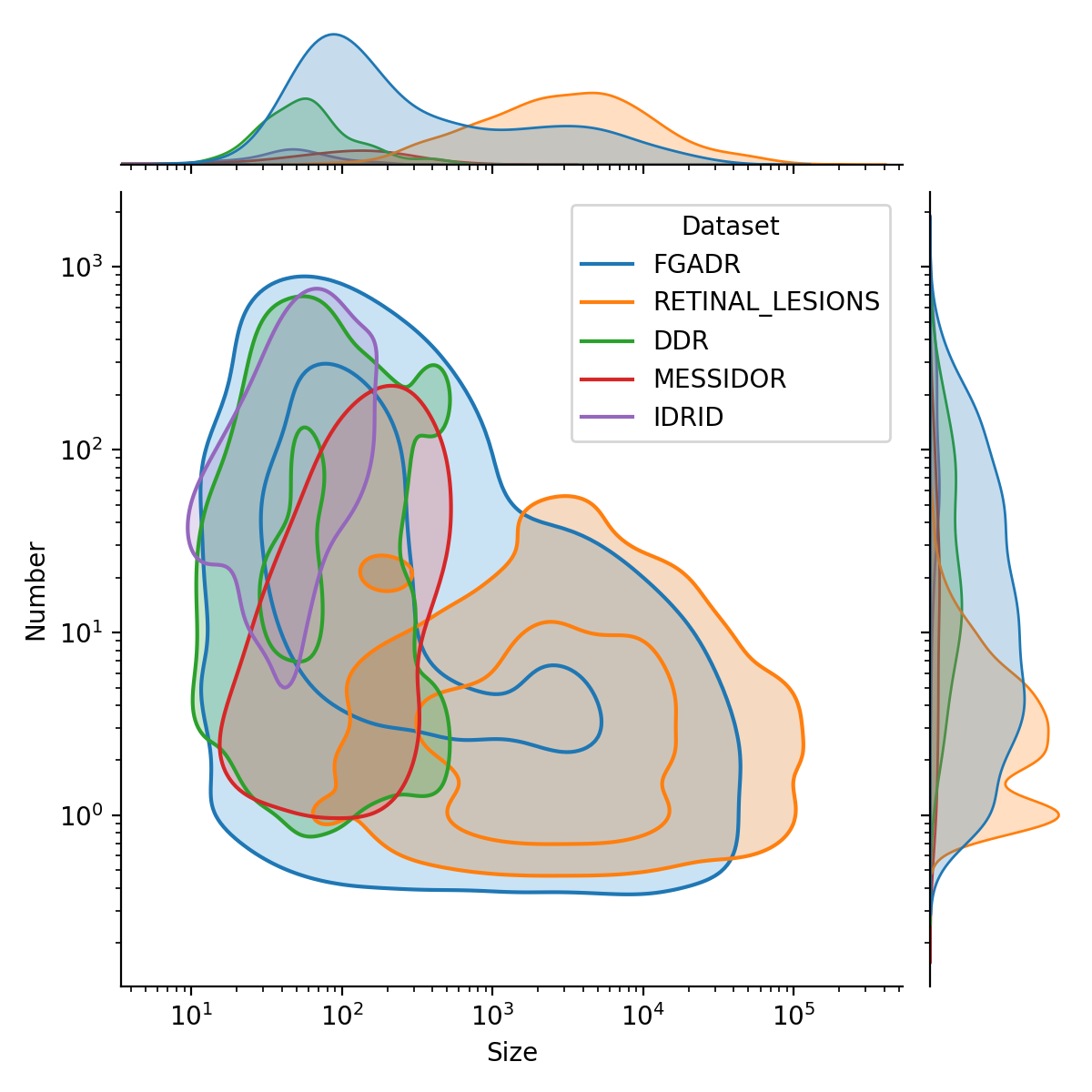}}
			\centerline{(a) Exudate distribution}\medskip
		\end{minipage}
		\hfill
		\begin{minipage}[b]{0.49\linewidth}
			\centering
			\centerline{\includegraphics[width=4.0cm]{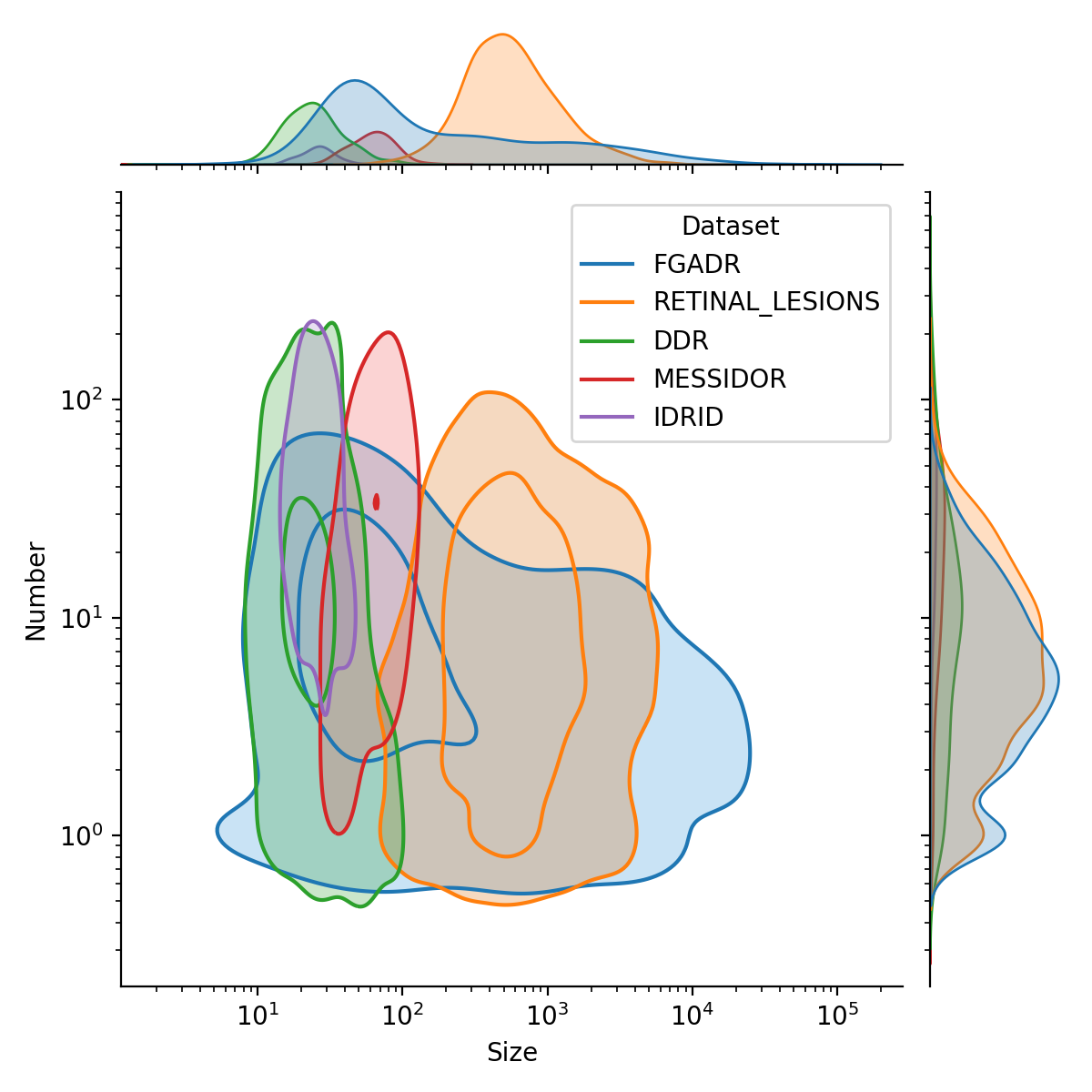}}
			\centerline{(b) Microaneurysm distribution}\medskip
		\end{minipage}
		\caption{Joint distribution of lesions size (x-axis) and count (y-axis) per image for each dataset. For visualization purposes, both axes are in log-scale. The lower-right corner corresponds to coarser labels and the upper-left corner to finer ones.
			The overlapping distributions reveal the existence of three clusters in labelling styles: coarse, fine, and mixed. 
		}
		\label{fig:joint_distribution}
	\end{figure}

	Datasets differ in size, annotated lesions (see Table \ref{tab:Datasets}), image quality and labelling protocol. To evaluate the quality, we used the Multiple Color-space Fusion Network (MCF-Net), developed by \cite{fuEvaluationRetinalImage2019}, which outputs a grade (Good, Usable, Reject) per image. Figure \ref{fig:image_quality} shows the results of image quality grading.
	To demonstrate the variability in terms of labelling style, we visualized the joint distribution of the lesion count and mean lesion area (in pixels) for every image in each dataset. Smaller and more numerous lesions tend to indicate a finer-grained labelling process. A clear separation between datasets appears for the annotation of exudates and microaneurysms (Figure \ref{fig:joint_distribution}). We thereby identified three clusters in terms of labelling styles: coarse for RET, fine-grained for IDRID, MESSIDOR and DDR, and a mix of coarse and fine-grained in the case of FGADR. 

	\subsection{Preprocessing}
	The images were cropped to completely remove the black background and fundus boundaries, 
	then resized and padded to reach a resolution of $1536\times 1536$, which was determined by hyperparameter tuning (described in Section \ref{sec:baseModel}). During training, random patches of size $512 \times 512$  were extracted from the full images. Data augmentation by random modification based on linear geometric transformations and HSV/brightness/contrast adjustment was applied.
	
	\subsection{Baseline model}
	\label{sec:baseModel}
	To conduct our generalization study, we first designed a baseline model and a training procedure that we followed for the rest of the experiments. We used Optuna \cite{akibaOptunaNextgenerationHyperparameter2019} to find the hyperparameters and the architecture that maximized the performance on a validation set. This tool was used to optimize the learning rate, optimizer, loss type and image resolution, as well as the network architecture by choosing its structure (UNet-like \cite{ronnebergerUNetConvolutionalNetworks2015a} and variants) and its encoder. The optimization was done over 1000 trials using the Tree-structured Parzen Estimator \cite{bergstraAlgorithmsHyperParameterOptimization} to suggest candidate parameters for each trial.
	
	\subsection{Dataset combinations}
	Each of the five datasets was split into training and testing sets. For IDRID and DDR, the division is already provided. For the three others, 30\% of the images were left out for testing and the rest served for training/validation (85\%-15\% respectively).
	We explored how combining the different training sets affects performance. Given $D=5$ datasets, we trained $N=31$ models (one for every possible combination:
	$
	N = \sum_{i=1}^{D} \binom{D}{i} = 2^D -1
	$).
	
	\subsection{Generalization strategies}
	\label{sec:generalization}
	A segmentation model is composed of an encoder $g_{\phi(t_{i}, {\theta_{j}})}$ and a decoder $f_{\psi(t_i, {\theta_{j}})}$, where $\phi(t_{i}, {\theta_{j}})$ (resp. $\psi(t_{i}, {\theta_{j}})$) represents the weights of the encoder (resp. decoder) after $t_i$ iterations of a training configured with the hyperparameters $\theta_{j}$. $t_i$ typically corresponds to the number of iterations at which a local minimum was reached on the validation loss. For ease of notation, we name the whole model $f_{\psi(t_i, {\theta_{j}})}$, but our experiments show that the decoder and encoder need to be considered separately.
	We compared the following generalization techniques:
	\begin{itemize}
		\item \textbf{Model ensemble:} as an approach to improve generalization, model ensembles can be built at inference time. Multiple trained models are used and their predictions are then simply averaged:
		\begin{equation}
		\tilde{y} = \frac{1}{n}\sum_{j}^{n} f_{\psi(t_i, {\theta_{j}})}(x)
		\end{equation}
		\item \textbf{Stochastic Weight Averaging (SWA):} this method averages the weights of a model obtained at different checkpoints during training. We refer to the original paper for further details \cite{izmailovAveragingWeightsLeads2018}.
		\begin{equation}
		\tilde{y} = f_{\frac{1}{n}\sum_{i}^{n} \psi(t_i, {\theta_{0}})}(x)
		\end{equation}
		
		\item \textbf{Model soup:} this recently proposed approach \cite{wortsmanModelSoupsAveraging} consists in averaging the weights of multiple models trained with slightly different hyperparameters; it was shown to improve classification performances.
		\begin{equation}
		\tilde{y} = f_{\frac{1}{n}\sum_{j}^{n} \psi(t_N, {\theta_{j}})}(x)
		\end{equation}
	\end{itemize}

	\section{Experiments}
	\subsection{Segmentation performance}
	The architecture tuning described in Section \ref{sec:baseModel} reached its peak performance for a U-Net using a ResNet-101 \cite{heDeepResidualLearning2016a} with Squeeze-and-Excitation blocks pretrained on ImageNet. We assessed its performance as shown in Table \ref{tab:SegmentationPerformance} compared to the published state-of-the-art, on three of the datasets. We adopted the evaluation protocol proposed in the IDRID challenge: for each lesion type, the area under the binned precision-recall curve (obtained for 11 segmentation thresholds) was computed.
	For each model in this table, the train and the test splits are derived from the same dataset.
	\begin{table}[h]
		\caption{Performance of our segmentation architecture using the area under the Precision-Recall curve.}\label{tab:SegmentationPerformance}
		\centering
		\begin{tabularx}{\columnwidth}{lXXXX|X}
			\hline
			\multicolumn{6}{c}{IDRID} \\
			\hline
			Models &  EX & HE &   MA & CWS & Mean\\
			\hline
			L-Seg \cite{guoLSegEndtoendUnified2019a} & 0.795 & \textbf{0.637} & 0.463 & 0.711 & \textbf{0.652} \\
			Deep-Bayesian \cite{garifullinDeepBayesianBaseline2021} & \textbf{0.842} & 0.593 & \textbf{0.484} & 0.641 & 0.640
			\\
			Ours & 0.825 & 0.620 & 0.451 & \textbf{0.713} & \textbf{0.652}\\
			\hline
			\multicolumn{6}{c}{DDR} \\
			\hline
			L-Seg \cite{guoLSegEndtoendUnified2019a} & 0.555 & 0.359 & 0.105 & 0.265 & 0.321\\
			Ours & \textbf{0.592} & \textbf{0.408} & \textbf{0.281} & \textbf{0.434} & \textbf{0.429}\\
			\hline
			\multicolumn{6}{c}{Retinal Lesions} \\
			\hline
			Lesion-Net-8s \cite{weiLearnSegmentRetinal2021a} & \textbf{0.740} & \textbf{0.617} & 0.377 & \textbf{0.575} & \textbf{0.577}\\
			Ours & 0.724 & 0.589 & \textbf{0.451} & 0.540 & 0.571\\
			\hline
		\end{tabularx}
	\end{table}
	Overall, the results indicate that our architecture performs quite closely to the state of the art, depending on the dataset and lesions considered.
	
	\subsection{Datasets combination}
	To quantify the segmentation performance, we used a single metric, namely the Dice score: $d(X, Y) = 2\frac{|X \cap Y|}{|X| + |Y|}$.
	
\begin{figure}[h]
		\begin{minipage}[b]{\columnwidth}
			\centerline{\includegraphics[width=\columnwidth]{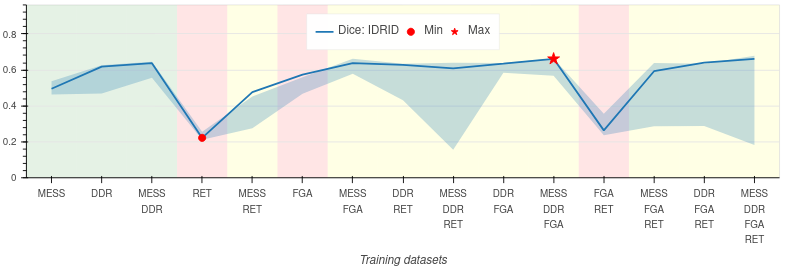}}
			\centerline{(a) IDRID-Test}\medskip
		\end{minipage}
		\begin{minipage}[b]{\columnwidth}
			\centerline{\includegraphics[width=\columnwidth]{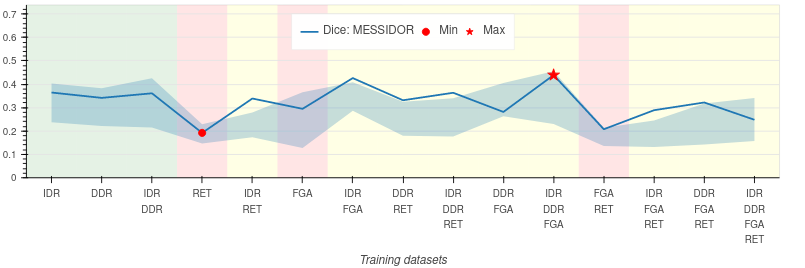}}
			\centerline{(b) MESSIDOR-Test}\medskip
		\end{minipage}
		\caption{ 
			Results of "leave one out" scenarios for different dataset combinations. The x-axis represents the training dataset combinations ordered by total number of images. In each case, the background colour corresponds to the overall labelling style (green=fine-grained, yellow=mixed, red=coarse). The shaded areas show the spread over 8 models trained in each case with different seeds. (a) Dice on IDRID: the best score (star) is obtained when training on a combination of datasets from overlapping fine-grained clusters (MES, DDR, FGA) (b) Dice on MES: the best score (star) is also obtained when training on fine-grained dataset 
			while the worst score (dot) is obtained when training on a coarse dataset (RET).}
		\label{fig:leaveOneOut}
	\end{figure}

	This metric was computed separately per test dataset. In Figure \ref{ref:generalizationTechniques}, an average across test sets was done to obtain a single Dice score.
	Initially, 31 models were trained, each with a different combination of training sets as described above. To analyse the results, we employed different "leave-one-out" scenarios, where one or more datasets were used for testing purposes and their training splits were excluded from the training sets. Figure \ref{fig:leaveOneOut} shows two of these scenarios. It illustrates the connection between the clusters observed in Figure \ref{fig:joint_distribution} (coarse vs. fine labels) and the influence from one dataset to another depending on whether they belong to the same cluster. Figure \ref{fig:leaveOneOut} indicates the spread in performances obtained using 8 models trained with slightly different hyperparameters. In particular, Figure \ref{fig:leaveOneOut} (a) shows how this spread can significantly increase when a combination of datasets from non-overlapping clusters is used for training. 
	The main results of the dataset combination experiments are summarized in Table \ref{tab:SegmentationPerformanceCombination}.
	\begin{table}[h]
		\caption{Segmentation performance (Dice score) based on the training sets combination.}\label{tab:SegmentationPerformanceCombination}
		\centering
		\begin{tabularx}{\columnwidth}{X|r}
			\hline
			\multicolumn{2}{c}{IDRID-Test} \\
			\hline
			With iid labels (MES+DDR) & 0.634 \\
			With only coarse (RET) & 0.222 \\
			With fine+coarse (MES+DDR+FGA) & \textbf{0.661} \\
			\hline
			\multicolumn{2}{c}{Messidor-Test} \\
			\hline
			With iid labels (IDR+DDR) & 0.422 \\
			With only coarse (RET) & 0.192 \\
			With fine+coarse (IDR+DDR+FGA) & \textbf{0.439} \\
			\hline
			\multicolumn{2}{c}{DDR-Test} \\
			\hline
			With iid labels (MES+IDR) & \textbf{0.426} \\
			With only coarse (RET) & 0.186 \\
			With fine+coarse (IDR+MES+FGA) & 0.409 \\
			\hline
		\end{tabularx}
	\end{table}
	We conclude that integrating coarsely annotated images within the training set can improve the test performance on a finely labelled dataset; indeed, this is the case when combining FGADR with fine-grained datasets. However, using RET mostly decreases the performance; this can be explained either by the lower quality of its images (see Figure \ref{fig:image_quality}) or by its annotation style being too different from the rest of the datasets.
	
	\subsection{Generalization Strategies}
	
\begin{figure}[h]
		\includegraphics[width=\columnwidth]{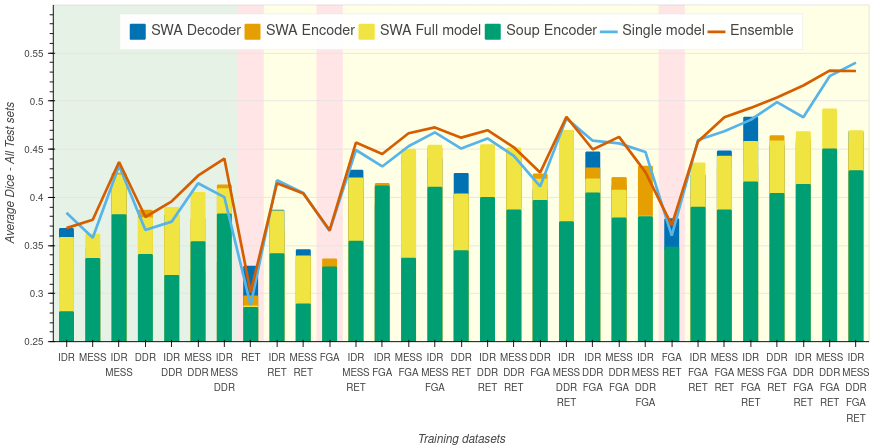}
		\centering
		\caption{Average Dice score for ensemble, SWA and model soup, for the 31 different training sets.}
		\label{ref:generalizationTechniques}
	\end{figure}

	We studied the performance obtained with the methods described in Section \ref{sec:generalization}. For SWA and model soup, which both average the weights of multiples models, we applied the technique either on the encoder, the decoder or the full model. Interestingly, for the model soup, combining the weights of the decoder or of the full-model simply did not work. Results are presented in Figure \ref{ref:generalizationTechniques}. In fact, model soups never outperformed the baseline model; SWA, either on the encoder or the full model, occasionally did so. 
	In particular, we notice its effectiveness on the IDR-MES-DDR combination, which all belong to the same cluster of annotation styles identified in Figure \ref{fig:joint_distribution}.
	As expected, the model ensemble approach improves the performance in most cases, but it is significantly more computationally demanding. In Figure \ref{ref:generalizationTechniques}, we averaged the predictions made by four distinct models.

	\section{Discussion and future work}
	By characterizing several publicly available datasets, we showed the existence of clusters in terms of retinal lesion annotation style, and we observed how this factor influences the behavior of the different models tested. We compared existing approaches aimed at achieving better generalization when trying to learn from multiple datasets. In so doing, we introduced model soups and SWA for the segmentation problem. However, these techniques did not provide the boosts in performance observed in their original papers, which were aimed at classification. One possible explanation is that the smaller amount of data available for segmentation (compared to the classification task) and the potentially 
	higher variability in terms of groundtruth result in a more irregular 
	loss function for which weights averaging is less effective. Further work will be required to explore the conditions of applicability of such techniques.
	Moreover, our study did not consider the literature related to noisy-labels learning, which aims at discovering the true distribution of a structure 
	by learning from multiple noisy sources. Similarly, we left unexplored the body of work on source-domain adaptation, which could compensate for the differences between the datasets, including in the label space. Finally, an important question remains regarding the effect of the lesion labelling style on disease classification and whether it influences the performance of a segmentation-guided classifier.

	\section{Compliance with ethical standards}
	This study used public data and ethical approval was not required as confirmed by the license attached with the data.
	
	\section{Acknowledgments}
	No funding was received for conducting this study. The authors have no relevant financial or non-financial interests to disclose.
	\bibliographystyle{IEEEbib}
	\bibliography{refs}

\begin{thebibliography}{10}

\bibitem{flaxelDiabeticRetinopathyPreferred2020}
C.~J. Flaxel, R.~A. Adelman, S.~T. Bailey, A.~Fawzi, J.~I. Lim, G.~A.
  Vemulakonda, and G.~Ying,
\newblock ``Diabetic {{Retinopathy Preferred Practice
  Pattern}}\textregistered,''
\newblock {\em Ophthalmology}, vol. 127, no. 1, pp. P66--P145, Jan. 2020.

\bibitem{lamWorldwideDiabetesEpidemic2012}
D.~W. Lam and D.~LeRoith,
\newblock ``The worldwide diabetes epidemic,''
\newblock {\em Current Opinion in Endocrinology, Diabetes and Obesity}, vol.
  19, no. 2, pp. 93--96, Apr. 2012.

\bibitem{kalogeropoulosRoleTeleophthalmologyDiabetic2020}
D.~Kalogeropoulos, C.~Kalogeropoulos, M.~Stefaniotou, and M.~Neofytou,
\newblock ``The role of tele-ophthalmology in diabetic retinopathy screening,''
\newblock {\em Journal of Optometry}, vol. 13, no. 4, pp. 262--268, Oct. 2020.

\bibitem{yanLearningMutuallyLocalGlobal2019a}
Z.~Yan, X.~Han, C.~Wang, Y.~Qiu, Z.~Xiong, and S.~Cui,
\newblock ``Learning {{Mutually Local-Global U-Nets For High-Resolution Retinal
  Lesion Segmentation In Fundus Images}},''
\newblock in {\em 2019 {{IEEE}} 16th {{International Symposium}} on
  {{Biomedical Imaging}} ({{ISBI}} 2019)}, Apr. 2019, pp. 597--600.

\bibitem{zhouBenchmarkStudyingDiabetic2021c}
Y.~Zhou, B.~Wang, L.~Huang, S.~Cui, and L.~Shao,
\newblock ``A {{Benchmark}} for {{Studying Diabetic Retinopathy}}:
  {{Segmentation}}, {{Grading}}, and {{Transferability}},''
\newblock {\em IEEE Transactions on Medical Imaging}, vol. 40, no. 3, pp.
  818--828, Mar. 2021.

\bibitem{weiLearnSegmentRetinal2021a}
Q.~Wei, X.~Li, W.~Yu, X.~Zhang, Y.~Zhang, B.~Hu, B.~Mo, D.~Gong, N.~Chen,
  D.~Ding, and Y.~Chen,
\newblock ``Learn to {{Segment Retinal Lesions}} and {{Beyond}},''
\newblock in {\em 2020 25th {{International Conference}} on {{Pattern
  Recognition}} ({{ICPR}})}. Jan. 2021, pp. 7403--7410, {IEEE Computer
  Society}.

\bibitem{guoLSegEndtoendUnified2019a}
S.~Guo, T.~Li, H.~Kang, N.~Li, Y.~Zhang, and K.~Wang,
\newblock ``L-{{Seg}}: {{An}} end-to-end unified framework for multi-lesion
  segmentation of fundus images,''
\newblock {\em Neurocomputing}, vol. 349, pp. 52--63, July 2019.

\bibitem{porwalIndianDiabeticRetinopathy2018a}
P.~Porwal,
\newblock ``Indian {{Diabetic Retinopathy Image Dataset}} ({{IDRiD}}),'' Apr.
  2018.

\bibitem{liDiagnosticAssessmentDeep2019b}
T.~Li, Y.~Gao, K.~Wang, S.~Guo, H.~Liu, and H.~Kang,
\newblock ``Diagnostic assessment of deep learning algorithms for diabetic
  retinopathy screening,''
\newblock {\em Information Sciences}, vol. 501, pp. 511--522, Oct. 2019.

\bibitem{izmailovAveragingWeightsLeads2018}
P.~Izmailov, D.~Podoprikhin, T.~Garipov, D.~Vetrov, and A.~G. Wilson,
\newblock ``Averaging weights leads to wider optima and better generalization:
  34th {{Conference}} on {{Uncertainty}} in {{Artificial Intelligence}} 2018,
  {{UAI}} 2018,''
\newblock pp. 876--885.

\bibitem{wortsmanModelSoupsAveraging}
M.~Wortsman, G.~Ilharco, S.~Y. Gadre, R.~Roelofs, R.~Gontijo-Lopes, S.~Morcos,
  H.~Namkoong, A.~Farhadi, Y.~Carmon, S.~Kornblith, and L.~Schmidt,
\newblock ``Model soups: averaging weights of multiple fine-tuned models
  improves accuracy without increasing inference time,''
\newblock in {\em Proceedings of the 39th International Conference on Machine
  Learning}. 17--23 Jul 2022, vol. 162 of {\em Proceedings of Machine Learning
  Research}, pp. 23965--23998, PMLR.

\bibitem{decenciere_feedback_2014}
E.~Decencière, X.~Zhang, G.~Cazuguel, B.~Lay, B.~Cochener, C.~Trone, P.~Gain,
  R.~Ordonez, P.~Massin, A.~Erginay, B.~Charton, and J.C Klein,
\newblock ``Feedback on a publicly distributed database: the messidor
  database,''
\newblock {\em Image Analysis \& Stereology}, vol. 33, no. 3, pp. 231--234,
  Aug. 2014.

\bibitem{maples_dr}
Gabriel Lepetit-Aimon, Clément Playout, Marie~Carole Boucher, Renaud Duval,
  Michael~H Brent, and Farida Cheriet,
\newblock ``Maples-dr: Messidor anatomical and pathological labels for
  explainable screening of diabetic retinopathy,''
\newblock 2024.

\bibitem{fuEvaluationRetinalImage2019}
H.~Fu, B.~Wang, J.~Shen, S.~Cui, Y.~Xu, J.~Liu, and L.~Shao,
\newblock ``Evaluation of {{Retinal Image Quality Assessment Networks}} in
  {{Different Color-Spaces}},''
\newblock in {\em Medical {{Image Computing}} and {{Computer Assisted
  Intervention}} \textendash{} {{MICCAI}} 2019: 22nd {{International
  Conference}}, {{Shenzhen}}, {{China}}, {{October}} 13\textendash 17, 2019,
  {{Proceedings}}, {{Part I}}}, {Berlin, Heidelberg}, Oct. 2019, pp. 48--56,
  {Springer-Verlag}.

\bibitem{akibaOptunaNextgenerationHyperparameter2019}
T.~Akiba, S.~Sano, T.~Yanase, T.~Ohta, and M.~Koyama,
\newblock ``Optuna: {{A Next-generation Hyperparameter Optimization
  Framework}},''
\newblock in {\em Proceedings of the 25th {{ACM SIGKDD International
  Conference}} on {{Knowledge Discovery}} \& {{Data Mining}}}, {New York, NY,
  USA}, July 2019, {{KDD}} '19, pp. 2623--2631, {Association for Computing
  Machinery}.

\bibitem{ronnebergerUNetConvolutionalNetworks2015a}
O.~Ronneberger, P.~Fischer, and T.~Brox,
\newblock ``U-{{Net}}: {{Convolutional Networks}} for {{Biomedical Image
  Segmentation}},''
\newblock in {\em Medical {{Image Computing}} and {{Computer-Assisted
  Intervention}} \textendash{} {{MICCAI}} 2015}, {Cham}, 2015, Lecture
  {{Notes}} in {{Computer Science}}, pp. 234--241, {Springer International
  Publishing}.

\bibitem{bergstraAlgorithmsHyperParameterOptimization}
J.~Bergstra, R.~Bardenet, Y.~Bengio, and B.~K\'{e}gl,
\newblock ``Algorithms for hyper-parameter optimization,''
\newblock in {\em Advances in Neural Information Processing Systems},
  J.~Shawe-Taylor, R.~Zemel, P.~Bartlett, F.~Pereira, and K.Q. Weinberger, Eds.
  2011, vol.~24, Curran Associates, Inc.

\bibitem{heDeepResidualLearning2016a}
K.~He, X.~Zhang, S.~Ren, and J.~Sun,
\newblock ``Deep {{Residual Learning}} for {{Image Recognition}},''
\newblock in {\em 2016 {{IEEE Conference}} on {{Computer Vision}} and {{Pattern
  Recognition}} ({{CVPR}})}, June 2016, pp. 770--778.

\bibitem{garifullinDeepBayesianBaseline2021}
A.~Garifullin, L.~Lensu, and H.~Uusitalo,
\newblock ``Deep {{Bayesian}} baseline for segmenting diabetic retinopathy
  lesions: {{Advances}} and challenges,''
\newblock {\em Computers in Biology and Medicine}, vol. 136, pp. 104725, Sept.
  2021.

\end{thebibliography}
	
\end{document}